\documentclass[10pt,twocolumn,letterpaper]{article}

\usepackage[pagenumbers]{cvpr} 

\usepackage[dvipsnames]{xcolor}

\usepackage{multicol}
\usepackage{multirow}
\usepackage{booktabs}
\usepackage{utfsym}
\usepackage{makecell}
\usepackage{caption}
\usepackage{subcaption}
\usepackage{appendix}

\usepackage{tikz,pgfplots,pgfplotstable}
\usepackage{color}
\usepackage{colortbl,hhline}
\usepgfplotslibrary[groupplots]
\definecolor{mediumelectricblue}{rgb}{0.12,0.314,0.588}
\definecolor{mossgreen2}{RGB}{138,154,91}
\definecolor{internationalorange}{RGB}{100,31,0}
\definecolor{lightsalmon}{RGB}{255,160,122}

\definecolor{selfyellow}{RGB}{255,209,0}
\definecolor{selfblue}{RGB}{89,138,234}
\definecolor{selfgreen}{RGB}{133,235,133}

\definecolor{cvprblue}{rgb}{0.21,0.49,0.74}
\usepackage[pagebackref,breaklinks,colorlinks,citecolor=cvprblue]{hyperref}

\def\paperID{} 
\def\confName{}
\def\confYear{}

\title{Forgery-aware Adaptive Transformer for Generalizable Synthetic\\Image Detection}

\author{
Huan Liu$^{1,3}$\thanks{Work done when H. Liu is a long-term intern at Baidu.}\quad
Zichang Tan$^{2}$\quad
Chuangchuang Tan$^{1,3}$\quad
Yunchao Wei$^{1,3}$\quad
Yao Zhao$^{1,3}$\quad
Jingdong Wang$^2$\\[1.2mm]
$^1$Institute of Information Science, Beijing Jiaotong University \quad
$^2$Baidu VIS\\
$^3$Beijing Key Laboratory of Advanced Information Science and Network Technology, Beijing, China\\
{\tt\small \{liu.huan,tanchuangchuang,yunchao.wei,yzhao\}@bjtu.edu.cn\quad \{tanzichang,wangjingdong\}@baidu.com}
}

\newlength\savewidth\newcommand\shline{\noalign{\global\savewidth\arrayrulewidth
  \global\arrayrulewidth 1pt}\hline\noalign{\global\arrayrulewidth\savewidth}}

\begin{document}
\maketitle

\begin{abstract}
In this paper, we study the problem of generalizable synthetic image detection, aiming to detect forgery images from diverse generative methods, e.g., GANs and diffusion models. 
Cutting-edge solutions start to explore the benefits of pre-trained models, and mainly follow the fixed paradigm of solely training an attached classifier, e.g., combining frozen CLIP-ViT with a learnable linear layer in UniFD \cite{ojha2023towards}.
However, our analysis shows that such a fixed paradigm is prone to yield detectors with insufficient learning regarding forgery representations.
We attribute the key challenge to the lack of forgery adaptation, and present a novel forgery-aware adaptive transformer approach, namely FatFormer.
Based on the pre-trained vision-language spaces of CLIP, FatFormer introduces two core designs for the adaption to build generalized forgery representations.
First, motivated by the fact that both image and frequency analysis are essential for synthetic image detection, we develop a forgery-aware adapter to adapt image features to discern and integrate local forgery traces within image and frequency domains.
Second, we find that considering the contrastive objectives between adapted image features and text prompt embeddings, a previously overlooked aspect, results in a nontrivial generalization improvement.
Accordingly, we introduce language-guided alignment to supervise the forgery adaptation with image and text prompts in FatFormer.
Experiments show that, by coupling these two designs, our approach tuned on 4-class ProGAN data attains a remarkable detection performance, achieving an average of $98\%$ accuracy to unseen GANs, and surprisingly generalizes to unseen diffusion models with $95\%$ accuracy.
\end{abstract}

\section{Introduction}
\label{sec:intro}
Recent years have witnessed the emergence and advancement of generative models, such as GANs \cite{goodfellow2014generative,karras2018progressive,karras2019style,karras2020analyzing} and diffusion models \cite{ho2020denoising,gu2022vector,nichol2022glide,dhariwal2021diffusion}. These models enable the creation of hyper-realistic synthetic images, thus raising the wide concerns of potential abuse and privacy threats. In response to such security issues, various forgery detection methods \cite{durall2020watch,jeong2022frepgan,jeong2022bihpf,shiohara2022detecting,shao2023detecting} have been developed, {\em e.g.}, image-based methods \cite{wang2020cnn,chai2020makes} focusing on low-level visual artifacts and frequency-based methods \cite{frank2020leveraging,qian2020thinking} relying on high-frequency pattern analysis. However, we observe big performance degradation when applying them to unseen images created by GANs or more recent diffusion models. How to address this problem has seen significant interest.

\begin{figure}[t]
\centering
\includegraphics[width=0.85\linewidth]{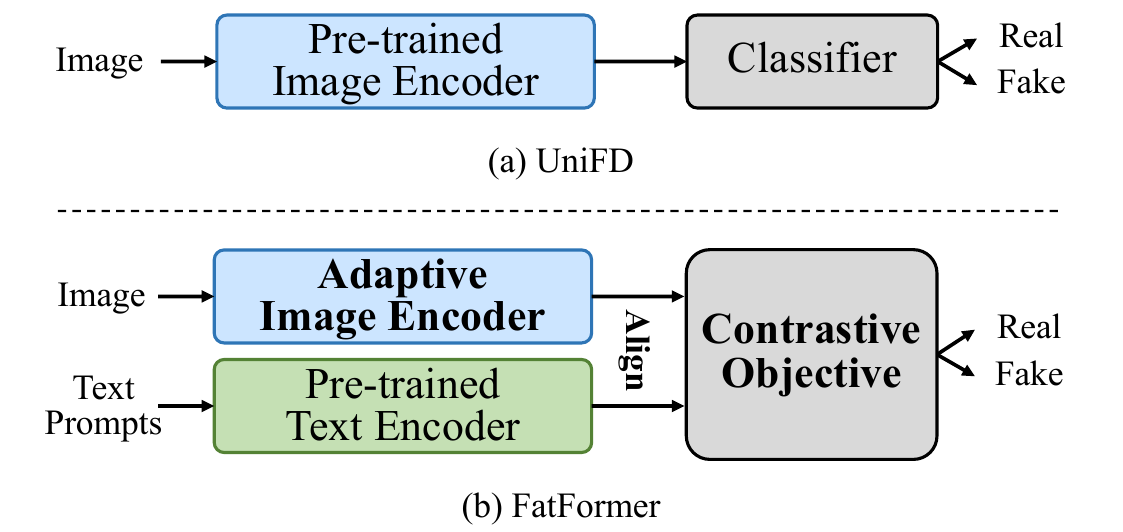}
\caption{\textbf{Comparison with fixed pre-trained paradigm.} Here, we illustrate the overview of UniFD~\cite{ojha2023towards} and our FatFormer. In contrast to training an attached classifier, FatFormer builds a forgery-aware adaptive transformer by aligning the representations of image and text prompts via contrastive objectives.}
\label{fig:figure1}
\vspace{-5mm}
\end{figure}

\begin{figure*}
\centering
\includegraphics[width=\linewidth]{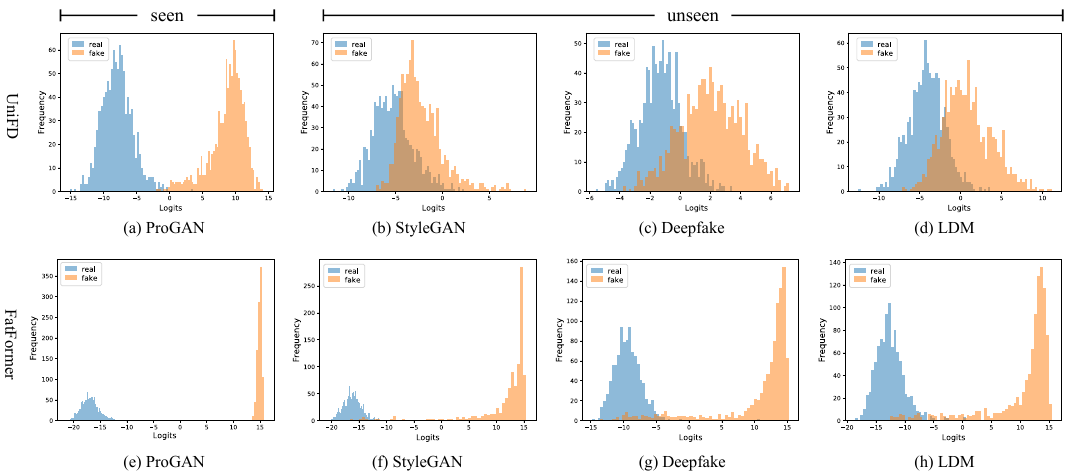}
\caption{\textbf{Logit distributions of extracted forgery features.} We compare the state-of-the-art UniFD~\cite{ojha2023towards} and our FatFormer with forgery adaptation, both tuned with 4-class ProGAN~\cite{karras2018progressive} data. A total of four testing GANs and diffusion models are considered, including ProGAN~\cite{karras2018progressive}, StyleGAN~\cite{karras2019style}, Deepfake~\cite{rossler2019faceforensics} and LDM~\cite{rombach2022high}, each randomly sampled $1$k real and $1$k fake images. Best view in color.}
\label{fig:figure2}
\vspace{-3mm}
\end{figure*}

Recent approaches \cite{tan2023learning,ojha2023towards} turn to explore the utilization of pre-trained models, following the fixed pre-trained paradigm of solely training an attached classifier, as shown in Figure~\ref{fig:figure1} (a). A notable example in this field is the UniFD proposed by Ojha~{\em et al.}~\cite{ojha2023towards}, where a pre-trained CLIP-ViT \cite{dosovitskiy2020image,radford2021learning} is employed to encode images into image features without learning. Subsequently, a linear layer is tuned as a classifier to determine the credibility of inputs.
At a very high level, their key to success is the employment of a pre-trained model in a frozen state, thus providing a learned universal representation (from the pre-training), yet not explicitly tuned in the current synthetic image detection task. 
In this way, such a representation will never be overfitted during training and thus preserves reasonable generalizability. However, we consider that such a frozen operation adopted by UniFD will also limit the capability of pre-trained models for learning strong and pertinent forgery features.

To verify our assumption, we qualitatively study the forgery discrimination of the fixed pre-trained paradigm by visualizing the logit distributions of UniFD \cite{ojha2023towards} across various generative models, as depicted in the top row of Figure~\ref{fig:figure2}.
The distribution reflects the degree of separation between `real' and `fake' during testing, thereby offering the extent of generalization of extracted forgery representations. 
One can see that there is a large overlap of `real' and `fake' regions when facing unseen GANs or diffusion models (Figure~\ref{fig:figure2} (b)-(d)), mistakenly, to identify these forgeries as `real' class. Moreover, even in the case of ProGAN~\cite{karras2018progressive} testing samples, which employ the identical generative model as the training data, the distinction between `read' and `fake' elements becomes increasingly indistinct (Figure~\ref{fig:figure2} (a) {\em vs.} (e)). 
We conclude that the fixed pre-trained paradigm is prone to yield detectors with insufficient learning regarding forgery artifacts, and attribute the key challenge to the lack of forgery adaptation that limits the full unleashing of potentials embedded in pre-trained models.

Driven by this analysis, we present a novel \textbf{F}orgery-aware \textbf{a}daptive \textbf{t}rans\textbf{Former} approach (Figure~\ref{fig:figure1} (b)), named FatFormer, for generalizable synthetic image detection.
In alignment with UniFD \cite{ojha2023towards}, FatFormer investigates CLIP \cite{radford2021learning} as the pre-trained model, which consists of a ViT \cite{dosovitskiy2020image} image encoder and a transformer \cite{vaswani2017attention} text encoder. Based on the pre-trained vision-language spaces of CLIP, our approach achieves the forgery adaptation by incorporating two core designs, ultimately obtaining well-generalized forgery representations with a distinct boundary between real and fake classes (Figure~\ref{fig:figure2} (e)-(h)).

First, motivated by the fact that both image and frequency domains are important for synthetic image detection, a forgery-aware adapter (FAA) is developed, comprising a pair of image and frequency forgery extractors. 
In the image domain, a lightweight convolution module is employed for extracting low-level forgery artifacts, such as blur textures and color mismatch \cite{li2020celeb}. 
On the other hand, for the frequency domain, we construct a grouped attention mechanism that dynamically aggregates frequency clues from different frequency bands of discrete wavelet transform (DWT) \cite{mallat1989theory}. 
By integrating these diverse forgery traces, FAA builds a comprehensive local viewpoint of image features essential for effective forgery adaptation.

\begin{figure*}[t]
\centering
\includegraphics[width=\textwidth]{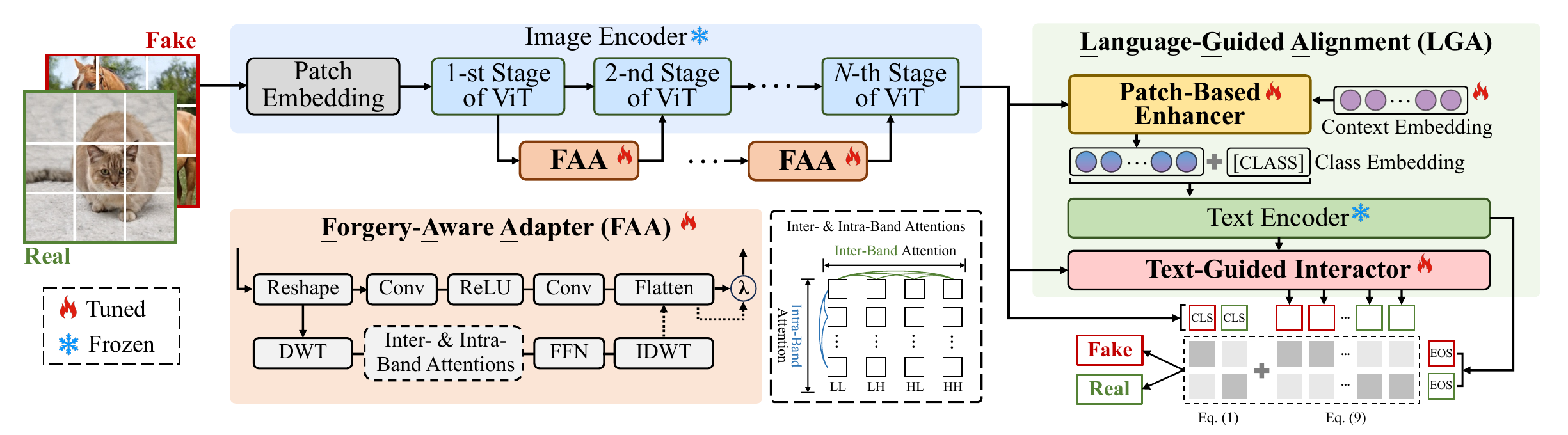}
\caption{\textbf{Our FatFormer architecture.} The ViT image encoder integrates forgery-aware adapters to effectively extract visual forgery features from input images. To supervise the forgery adaptation process, language-guided alignment is introduced. Specifically, taking two input images for example, we maximize the cosine similarities between paired (dark gray squares) image features and text prompt embeddings, while minimizing the unpaired ones (light gray squares). In testing, only the test image is required to calculate the forgery probability via a softmax of these similarities. Squared `CLS' and `EOS' represent the image CLS tokens and text prompt embeddings.} 
\label{fig:figure3}
\vspace{-4mm}
\end{figure*}

Second, instead of utilizing the binary cross-entropy loss applied to image features, we consider the contrastive objectives between image and text prompts, a previously overlooked aspect. This novel direction is inspired by the natural language supervision in CLIP-ViT's pre-training, typically more robust to overfitting by optimizing the similarity between image features and text prompt embeddings \cite{hu2021well}. 
Accordingly, language-guided alignment (LGA) is proposed, which encompasses a patch-based enhancer, designed to enrich the contextual relevance of text prompts by conditioning them on image patch tokens, as well as a text-guided interactor, that serves to align local image patch tokens with global text prompt embeddings, thereby directing the image encoder to concentrate on forgery-related representations.
Empirical results show that the forgery adaptation supervised by LGA obtains more generalized forgery representations, thus improving the generalizability of synthetic image detection. 

Our adaptive approach FatFormer significantly outperforms recent methods with the fixed pre-trained paradigm. 
Notably, we achieve $98.4\%$ ACC and $99.7\%$ AP on $8$ types of GANs, and $95.0\%$ ACC and $98.8\%$ AP on $10$ types of unseen diffusion images, using limited ProGAN training data. 
We hope our findings can facilitate the development of pre-trained paradigms in this field.

\section{Related Work}
\noindent \textbf{Synthetic image detecting.}
Due to the increasing concerns about generative models, many works are proposed to address the problem of synthetic image detection, which can be roughly divided into image-based methods \cite{marra2019gans,rossler2019faceforensics,zhao2021learning,wang2023dire}, frequency-based methods \cite{frank2020leveraging,jeong2022bihpf,jeong2022frepgan}, and pre-trained-based methods \cite{tan2023learning,ojha2023towards}.
For instance, Yu {\em et al.} \cite{yu2019attributing} find images generated by GANs have unique fingerprints, which can be utilized as forgery traces for detection. Wang {\em et al.} \cite{wang2020cnn} adopt various data augmentations and large-scale GAN images to improve the generalization to unseen testing data. Qian {\em et al.} \cite{qian2020thinking} introduce frequency analysis into the detection framework, using local frequency statistics and decomposed high-frequency components for forgery detection. 
More recently, many works have focused on the fixed pre-trained paradigm of freezing the pre-trained model and adopting an attached classifier for forgery detection. 
For example, Lgrad \cite{tan2023learning} turns the detection problem into a pre-trained-based transformation-dependent problem, and utilizes gradient features from the frozen pre-trained model as forgery cues. 
Furthermore, Ojha {\em et al.} \cite{ojha2023towards} propose UniFD to explore the potential of the vision-language model, {\em i.e.}, CLIP \cite{radford2021learning}, for synthetic image detection. They observe that training a deep network fails to detect fake images from new breeds and employs the frozen CLIP-ViT \cite{dosovitskiy2020image,radford2021learning} to extract forgery features, followed by a linear classifier.

In this paper, our motivation is different from the closely-related approach UniFD \cite{ojha2023towards}. UniFD attempts to adopt a frozen pre-trained model to extract forgery representations `without learning'. In contrast, our approach aims to demonstrate that the forgery adaptation of pre-trained models is essential for the generalizability of synthetic image detection.

\noindent \textbf{Efficient transfer learning.}
The latest progress in transfer learning shows the potential for efficient fine-tuning of pre-trained models, especially in the NLP field. Unlike traditional strategies, such as linear-probing \cite{he2022masked} and full fine-tuning \cite{zhuang2021comprehensive}, efficient transfer learning only adds learnable modules with a few parameters, such as prompt learning \cite{lester2021power} and adapter-based methods \cite{houlsby2019parameter,hu2021lora}.
Inspired by this, many efficient transfer learning works are proposed for vision \cite{jia2022visual,chen2022adaptformer} and vision-language models \cite{zhou2022conditional,zhou2022learning}.
Unlike UniFD \cite{ojha2023towards} with linear probing, this paper investigates the efficient transfer learning for generalizable synthetic image detection and first proposes an adaptive transformer with contrastive objectives.

\section{FatFormer}
\subsection{Overview}
The overall structure of FatFormer is illustrated in Figure~\ref{fig:figure3}. FatFormer is composed of two pre-trained encoders for both image and text prompts, as well as the proposed forgery-aware adapter (Section~\ref{sec:FAA}) and language-guided alignment (Section~\ref{sec:LGA}). 
This framework predicts the forgery probability by calculating the softmax of cosine similarities between image features and text prompt embeddings.

\noindent \textbf{Vanilla CLIP.} Following UniFD \cite{ojha2023towards}, we adopt CLIP \cite{radford2021learning} as the pre-trained model with a ViT \cite{dosovitskiy2020image} image encoder and transformer \cite{vaswani2017attention} text encoder, respectively. Given an image $x\in \mathbb{R}^{3\times H \times W}$, with height $H$ and width $W$, CLIP converts it into a $D$-dimensional image features $f_{img}\in \mathbb{R}^{(1+N) \times D}$, where $1$ represents the image CLS token, $N=\frac{HW}{P^2}$ denotes the image patch tokens and $P$ is the patch size. Meanwhile, the text encoder takes language text $t$ and generates the text prompt embeddings $f_{text} \in \mathbb{R}^{M \times D}$ from the appended EOS tokens in the text encoder, where $M$ denotes the number of classes (in this paper, $M=2$). Two encoders are jointly trained to optimize the cosine similarity between the image CLS token and text prompt embeddings using contrastive loss. After pre-training, we can utilize the re-assembled text descriptions for zero-shot testing, {\em e.g.}, a simple template of `this photo is [CLASS]', where `[CLASS]' is replaced by class names like `real' or `fake'. Given the testing image and text prompts, we have the predicted similarity of class $i \in \{0, 1\}$, where $0$ represents `real' and $1$ is `fake', as follows
\begin{equation}
    S(i) = {\rm cos}(f_{img}^{(0)}, f_{text}^{(i)}), \label{eq:vanilla_clip}
\end{equation}
where ${\rm cos}(\cdot)$ is the cosine similarity, $f_{img}^{(0)}$ denotes the image CLS token at index $0$ of $f_{img}$. Further, the corresponding possibility can be derived via a softmax function
\begin{equation}
    P(i) = \frac{{\rm exp}(S(i) / \tau)}{\sum_{k}{\rm exp}(S(k) / \tau)},
\end{equation}
where $\tau$ is the temperature parameter.

\subsection{Forgery-aware adapter (FAA)}
\label{sec:FAA}
To adapt the image features for effective forgery adaptation, we insert forgery-aware adapters to bridge adjacent ViT stages, each encompassing multiple ViT layers, in the image encoder, as shown in Figure~\ref{fig:figure3}. 
These adapters discern and integrate forgery traces within both image and frequency domains, enabling a comprehensive local viewpoint of image features.

\noindent \textbf{Image forgery extractor.}
In the image domain, FAA constructs a lightweight image forgery extractor, comprising two convolution layers and a ReLU layer for capturing low-level image artifacts, as follows
\begin{align}
    \hat{g}_{img}^{(j)} = {\rm Conv}({\rm ReLU}({\rm Conv}(g_{img}^{(j)}))),
\end{align}
where $\hat{g}_{img}^{(j)}$ represents the adapted forgery-aware image features from FAA in $j$-th ViT stage, and $g_{img}^{(j)}$ is the vanilla features from the last multi-head attention module in $j$-th ViT stage. Here, we omit the reshape operators.

\noindent \textbf{Frequency forgery extractor.}
For the frequency domain, a grouped attention mechanism is proposed to mine forgery traces in the frequency bands of discrete wavelet transform (DWT) \cite{mallat1989theory}.
Although previous detection methods \cite{jeong2022frepgan,qian2020thinking} adopt fast Fourier transform \cite{brigham1967fast} and discrete cosine transform \cite{rao2014discrete}, they destroy the position information \cite{li2020wavelet} in the transformed frequency domain, which is crucial in the context of attention modeling \cite{dosovitskiy2020image}. 
Thus, we utilize DWT as the transform function, retaining the spatial structure of image features, which decomposes the inputs into $4$ distinct frequency bands, including LL, LH, HL, and HH. 
Here, combinations of `L' and `H' represent the combined low and high pass filters.
Then, two grouped attention modules, {\em i.e.}, inter-band attention and intra-band attention, are proposed for the extraction of frequency clues. 
As indicated in Figure~\ref{fig:figure3}, the inter-band attention explicitly explores the interactions across diverse frequency bands, while the intra-band attention builds interactions within each frequency band.
This design achieves the dynamical aggregation of different positions and bands, rather than manual weighting like F3Net \cite{qian2020thinking}.
In practice, we implement them with multi-head attention modules \cite{vaswani2017attention}. Finally, FFN and inverse discrete wavelet transform (IDWT) are used to obtain forgery-aware frequency features $\hat{g}_{freq}^{(j)}$, which are transformed back into the image domain for further incorporation.

To prevent introducing hyper-parameters, we leverage a learnable scale factor $\lambda$ to control the information from image and frequency domains as the final adapted image features of $j$-th stage of ViT, which will be sent to the first multi-head attention module in the next $(j+1)$-th stage.
\begin{equation}
    \hat{g}^{(j)} = \hat{g}_{img}^{(j)} + \lambda \cdot \hat{g}_{freq}^{(j)}.
\end{equation}

\subsection{Language-guided alignment (LGA)}
\label{sec:LGA}
To supervise the forgery adaptation of FatFormer, language-guided alignment is proposed by considering the contrastive objectives between image and text prompts. 
In a bit more detail, LGA has a patch-based enhancer that enriches the context of text prompts, and a text-guided interactor that aligns the local image patch tokens with global text prompt embeddings. 
Finally, we implement an augmented contrastive objective for the loss calculation.

\noindent \textbf{Patch-based enhancer.}
Instead of using hand-crafted templates as prompts, FatFormer has a soft prompt design by adopting auto context embeddings, following \cite{zhou2022conditional,zhou2022learning}. 
Since synthetic image detection relies on local forgery details \cite{zhao2021multi,chen2021local}, we develop a patch-based enhancer to enhance the contextual relevance of prompts via the condition of local image patch tokens, deriving forgery-relevant prompts context. Specifically, we first compute the image patch tokens $f_{img}^{(1:N)} \in \mathbb{R}^{N\times D}$ in the image encoder. Then, given $C$ context embeddings $p_{ctx}\in \mathbb{R}^{C\times D}$, we have
\begin{align}
    A_{pbe} = p_{ctx} \cdot (f_{img}^{(1:N)})^T,
\end{align}
where $A_{pbe} \in \mathbb{R}^{C\times N}$ is the similarity matrix in patch-based enhancer. We use $A_{pbe}$ to represent the intensity of image patch tokens for constructing each context embedding, as follows
\begin{align}
    \hat{p}_{ctx} = {\rm softmax}(A_{pbe}) \cdot f_{img}^{(1:N)} + p_{ctx}. \label{eq:ctx}
\end{align}
Finally, we can obtain the set of possible text prompts by combining the enhanced context $\hat{p}_{ctx}$ and $M$ [CLASS] embeddings, and send them to the text encoder.

\noindent \textbf{Text-guided interactor.}
To guide the image encoder focusing on forgery-related representation, we propose a text-guided interactor, which aligns the local image patch tokens with global text prompt embeddings. 
Specifically, given the text prompt embeddings $f_{text}$ from text encoder and image patch tokens $f_{img}^{(1:N)}$, our text-guided interactor calculates the similarity $A_{tgi}$ between them by
\begin{align}
    A_{tgi} =  f_{img}^{(1:N)} \cdot (f_{text})^T.
\end{align}
Similar to Eq.~(\ref{eq:ctx}), with $A_{tgi}$, sized $\mathbb{R}^{N\times M}$, we align the image patch tokens with text prompt embeddings by adaptively augmenting text representations, as follows
\begin{align}
    \hat{f}_{img}^{(1:N)} = {\rm softmax}(A_{tgi}) \cdot f_{text} + f_{img}^{(1:N)},
\end{align}
where $\hat{f}_{img}^{(1:N)}$ denotes the aligned image patch tokens. Together with the augmented contrastive objectives, the image encoder is guided to concentrate on forgery-related representation within each distinct image patch.

\noindent \textbf{Augmented contrastive objectives.} For the loss calculation, we consider augmented contrastive objectives that comprise two elements. 
The first is the cosine similarity in Eq.~(\ref{eq:vanilla_clip}) same as the vanilla CLIP. 
The second is the similarity between text prompt embeddings and aligned image patch tokens $\hat{f}_{img}^{(1:N)}$. With $t \in [1, N]$ and $i \in \{0, 1\}$, we have
\begin{align}
    S'(i) = \frac{1}{N}\sum_{t}{\rm cos}(\hat{f}_{img}^{(t)}, f_{text}^{(i)}).
    \label{eq:our_clip}
\end{align}
By merging similarities from Eq.~(\ref{eq:vanilla_clip}) and Eq.~(\ref{eq:our_clip}), our FatFormer describes a augmented probability $\hat{P}(i)$ by a softmax function, as follows
\begin{align}
    \hat{P}(i) &= \frac{{\rm exp}((S(i) + S'(i)) / \tau)}{\sum_{k} {\rm exp}((S(k) + S'(k)) / \tau)}. \label{eq:loss}
\end{align}
In practice, we apply the cross-entropy function on Eq.~(\ref{eq:loss}) with label $y \in \{0,1\}$ to calculate contrastive loss like the origin CLIP, as follows
\begin{align}
    \mathcal{L} = - y \cdot \log \hat{P}(y)-(1-y) \cdot \log (1-\hat{P}(y)).
\end{align}
\section{Experiments}
\subsection{Settings}
\noindent \textbf{Datasets.}
As generative methods are always coming up, we follow the standard protocol \cite{wang2020cnn,ojha2023towards,tan2023learning} that limits the accessible training data to only one generative model, while testing on unseen data, such as synthetic images from other GANs and diffusion models. Specifically, we train FatFormer on the images generated by ProGAN \cite{karras2018progressive} with two different settings, including 2-class (chair, horse) and 4-class (car, cat, chair, horse) data from \cite{wang2020cnn}. For evaluation, we collect the testing GANs dataset provided in \cite{wang2020cnn} and diffusion model datasets in \cite{ojha2023towards,wang2023dire}, which contain synthetic images and the corresponding real images. The testing GANs dataset includes ProGAN \cite{karras2018progressive}, StyleGAN \cite{karras2019style}, StyleGAN2 \cite{karras2020analyzing}, BigGAN \cite{brock2018large}, CycleGAN \cite{zhu2017unpaired}, StarGAN \cite{choi2018stargan}, GauGAN \cite{park2019semantic} and DeepFake \cite{rossler2019faceforensics}. On the other hand, the diffusion part consists of PNDM \cite{liu2021pseudo}, Guided \cite{dhariwal2021diffusion}, DALL-E \cite{ramesh2021zero}, VQ-Diffusion \cite{gu2022vector}, LDM \cite{rombach2022high}, and Glide \cite{nichol2022glide}. For LDM and Glide, we also consider their variants with different generating settings. More details can be found in their official papers.

\noindent \textbf{Evaluation metric.}
The accuracy (ACC) and average precision (AP) are reported as the main metrics during evaluation for each generative model, following the standard process \cite{wang2020cnn,ojha2023towards,tan2023learning}. To better evaluate the overall model performance over the GANs and diffusion model datasets, we also adopt the mean of ACC and AP on each dataset, denoted as ACC$_M$ and AP$_M$. 

\noindent \textbf{Implementation details.}
Our main training and testing settings follow the previous study \cite{ojha2023towards}. The input images are first resized into $256 \times 256$, and then image cropping is adopted to derive the final resolution of $224 \times 224$. We apply random cropping and random horizontal flipping at training, while center cropping at testing, both with no other augmentations. The Adam \cite{kingma2014adam} is utilized with betas of $(0.9, 0.999)$. We set the initial learning rate as $4 \times 10^{-4}$, training epochs as $25$, and adopt a total batch size of $256$. Besides, a learning rate schedule is used, decaying at every $10$ epochs by a factor of $0.9$.

\subsection{Main results}
This paper aims to build a better paradigm with pre-trained models for synthetic image detection. Therefore, we mainly compare our FatFormer with previous methods that adopt the fixed pre-trained paradigm, such as LGrad \cite{tan2023learning} and UniFD \cite{ojha2023towards}. In addition, to show the effectiveness of our approach, we also consider comparisons with existing image-based \cite{wang2020cnn,chai2020makes,shiohara2022detecting} and frequency-based methods \cite{durall2020watch,frank2020leveraging,qian2020thinking,jeong2022bihpf,jeong2022frepgan}.

\begin{table*}[!ht]
\centering
\renewcommand{\arraystretch}{1.3}
\caption{\textbf{Accuracy and average precision comparisons with state-of-the-art methods on GANs dataset.} We report the performance (in the formulation of ACC / AP) with two different training settings, including supervision from 2-class and 4-class ProGAN data, following \cite{wang2020cnn}. Besides, we also provide the reference (Ref) for previous frameworks. $\dagger$ denotes only trained on self-blended images of FF++ \cite{rossler2019faceforensics}. The performance (ACC$_M$ / AP$_M$) over the entire dataset is marked in \colorbox{lightgray!20}{gray}. The \textbf{best results} are highlighted in \textbf{bold}.}
\resizebox{\textwidth}{!}{
\begin{tabular}{c|l|c|cccccccc|c}
\multicolumn{2}{c|}{Methods} & Ref & ProGAN& StyleGAN& StyleGAN2 & BigGAN & CycleGAN & StarGAN & GauGAN & Deepfake & \cellcolor{lightgray!20}{Mean} \\
\shline
\multirow{9}{*}{\rotatebox{90}{2-class supervision}} 
& Wang \cite{wang2020cnn}          & CVPR 2020 & $64.6$ / $92.7 $ & $52.8$ / $82.8$ & $75.7$ / $96.6$ & $51.6$ / $70.5$ & $58.6$ / $81.5$ & $51.2 $ / $74.3 $ & $53.6$ / $86.6 $ & $50.6$ / $51.5$ & \cellcolor{lightgray!20}{$57.3$ / $79.6$} \\
& Durall \cite{durall2020watch}    & CVPR 2020 & $79.0$ / $73.9 $ & $63.6$ / $58.8$ & $67.3$ / $62.1$ & $69.5$ / $62.9$ & $65.4$ / $60.8$ & $99.4 $ / $99.4 $ & $67.0$ / $63.0 $ & $50.5$ / $50.2$ & \cellcolor{lightgray!20}{$70.2$ / $66.4$} \\
& Frank \cite{frank2020leveraging} & ICML 2020 & $85.7$ / $81.3 $ & $73.1$ / $68.5$ & $75.0$ / $70.9$ & $76.9$ / $70.8$ & $86.5$ / $80.8$ & $85.0 $ / $77.0 $ & $67.3$ / $65.3 $ & $50.1$ / $55.3$ & \cellcolor{lightgray!20}{$75.0$ / $71.2$} \\ 
& F3Net \cite{qian2020thinking}    & ECCV 2020 & $97.9$ / $100.0$ & $84.5$ / $99.5$ & $82.2$ / $99.8$ & $65.5$ / $73.4$ & $81.2$ / $89.7$ & $100.0$ / $100.0$ & $57.0$ / $59.2 $ & $59.9$ / $83.0$ & \cellcolor{lightgray!20}{$78.5$ / $88.1$} \\
& BiHPF \cite{jeong2022bihpf}      & WACV 2022 & $87.4$ / $87.4 $ & $71.6$ / $74.1$ & $77.0$ / $81.1$ & $82.6$ / $80.6$ & $86.0$ / $86.6$ & $93.8 $ / $80.8 $ & $75.3$ / $88.2 $ & $53.7$ / $54.0$ & \cellcolor{lightgray!20}{$78.4$ / $79.1$} \\
& FrePGAN \cite{jeong2022frepgan}  & AAAI 2022 & $99.0$ / $99.9 $ & $80.8$ / $92.0$ & $72.2$ / $94.0$ & $66.0$ / $61.8$ & $69.1$ / $70.3$ & $98.5 $ / $100.0$ & $53.1$ / $51.0 $ & $62.2$ / $80.6$ & \cellcolor{lightgray!20}{$75.1$ / $81.2$} \\
& LGrad \cite{tan2023learning}     & CVPR 2023 & $99.8$ / $100.0$ & $\bf 94.8$ / $\bf 99.7$ & $\bf 92.4$ / $\bf 99.6$ & $82.5$ / $92.4$ & $85.9$ / $94.7$ & $99.7 $ / $99.9 $ & $73.7$ / $83.2 $ & $60.6$ / $67.8$ & \cellcolor{lightgray!20}{$86.2$ / $92.2$} \\
& UniFD \cite{ojha2023towards}      & CVPR 2023 & $99.7$ / $100.0$ & $78.8$ / $97.4$ & $75.4$ / $96.7$ & $91.2$ / $99.0$ & $91.9$ / $99.8$ & $96.3 $ / $99.9 $ & $91.9$ / $100.0$ & $80.0$ / $89.4$ & \cellcolor{lightgray!20}{$88.1$ / $97.8$} \\
\hhline{~-----------}
& Ours                             &    $-$    & $\bf 99.8$ / $\bf 100.0$ & $87.7$ / $97.4$ & $91.1$ / $99.3$ & $\bf 98.9$ / $\bf 99.9$ & $\bf 99.9$ / $\bf 100.0$ & $\bf 100.0$ / $\bf 100.0$ & $\bf 99.9$ / $\bf 100.0$ & $\bf 89.4$ / $\bf 97.3$ & \cellcolor{lightgray!20}{$\bf 95.8$ / $\bf 99.2$} \\
\shline
\multirow{10}{*}{\rotatebox{90}{4-class supervision}} 
& Wang \cite{wang2020cnn}          & CVPR 2020 & $91.4$ / $99.4 $ & $63.8$ / $91.4$ & $76.4$ / $97.5$ & $52.9$ / $73.3$ & $72.7$ / $88.6$ & $63.8 $ / $90.8 $ & $63.9$ / $92.2$ & $51.7$ / $62.3$ & \cellcolor{lightgray!20}{$67.1$ / $86.9$} \\
& Durall \cite{durall2020watch}    & CVPR 2020 & $81.1$ / $74.4 $ & $54.4$ / $52.6$ & $66.8$ / $62.0$ & $60.1$ / $56.3$ & $69.0$ / $64.0$ & $98.1 $ / $98.1 $ & $61.9$ / $57.4$ & $50.2$ / $50.0$ & \cellcolor{lightgray!20}{$67.7$ / $64.4$} \\
& Frank \cite{frank2020leveraging} & ICML 2020 & $90.3$ / $85.2 $ & $74.5$ / $72.0$ & $73.1$ / $71.4$ & $88.7$ / $86.0$ & $75.5$ / $71.2$ & $99.5 $ / $99.5 $ & $69.2$ / $77.4$ & $60.7$ / $49.1$ & \cellcolor{lightgray!20}{$78.9$ / $76.5$} \\
& PatchFor \cite{chai2020makes}    & ECCV 2020 & $97.8$ / $100.0$ & $82.6$ / $93.1$ & $83.6$ / $98.5$ & $64.7$ / $69.5$ & $74.5$ / $87.2$ & $\bf 100.0$ / $100.0$ & $57.2$ / $55.4$ & $85.0$ / $93.2$ & \cellcolor{lightgray!20}{$80.7$ / $87.1$} \\
& F3Net \cite{qian2020thinking}    & ECCV 2020 & $99.4$ / $100.0$ & $92.6$ / $99.7$ & $88.0$ / $99.8$ & $65.3$ / $69.9$ & $76.4$ / $84.3$ & $\bf 100.0$ / $100.0$ & $58.1$ / $56.7$ & $63.5$ / $78.8$ & \cellcolor{lightgray!20}{$80.4$ / $86.2$} \\
& Blend$\dagger$ \cite{shiohara2022detecting} & CVPR 2022 & $58.8$ / $65.2$ & $50.1$ / $47.7$ & $48.6$ / $47.4$ & $51.1$ /	$51.9$ & $59.2$ / $65.3$ & $74.5$ / $89.2$ & $59.2$ / $65.5$ & $93.8$ / $99.3$ & \cellcolor{lightgray!20}{$61.9$ / $66.4$} \\
& BiHPF \cite{jeong2022bihpf}      & WACV 2022 & $90.7$ / $86.2 $ & $76.9$ / $75.1$ & $76.2$ / $74.7$ & $84.9$ / $81.7$ & $81.9$ / $78.9$ & $94.4 $ / $94.4 $ & $69.5$ / $78.1$ & $54.4$ / $54.6$ & \cellcolor{lightgray!20}{$78.6$ / $77.9$} \\
& FrePGAN \cite{jeong2022frepgan}  & AAAI 2022 & $99.0$ / $99.9 $ & $80.7$ / $89.6$ & $84.1$ / $98.6$ & $69.2$ / $71.1$ & $71.1$ / $74.4$ & $99.9$ / $100.0$ & $60.3$ / $71.7$ & $70.9$ / $91.9$ & \cellcolor{lightgray!20}{$79.4$ / $87.2$} \\
& LGrad \cite{tan2023learning}     & CVPR 2023 & $99.9$ / $100.0$ & $94.8$ / $\bf 99.9$ & $96.0$ / $99.9$ & $82.9$ / $90.7$ & $85.3$ / $94.0$ & $99.6 $ / $100.0$ & $72.4$ / $79.3$ & $58.0$ / $67.9$ & \cellcolor{lightgray!20}{$86.1$ / $91.5$} \\
& UniFD \cite{ojha2023towards}      & CVPR 2023 & $99.7$ / $100.0$ & $89.0$ / $98.7$ & $83.9$ / $98.4$ & $90.5$ / $99.1$ & $87.9$ / $99.8$ & $91.4 $ / $100.0$ & $89.9$ / $100.0$ & $80.2$ / $90.2$ & \cellcolor{lightgray!20}{$89.1$ / $98.3$} \\
\hhline{~-----------}
& Ours                             &    $-$    & $\bf99.9$ / $\bf100.0$& $\bf97.2$ / 99.8 & $\bf98.8$ / $\bf99.9$ & $\bf99.5$ / $\bf100.0$& $\bf99.3$ / $\bf100.0$& 99.8 / $\bf100.0$& $\bf99.4$ / $\bf100.0$& $\bf93.2$ / $\bf98.0$ & \cellcolor{lightgray!20}{$\bf98.4$ / $\bf99.7$} \\
\end{tabular}}
\label{tab:table1}
\end{table*}
\begin{table*}[!ht]
\centering
\renewcommand{\arraystretch}{1.3}
\caption{\textbf{Accuracy and average precision comparisons with state-of-the-art methods on diffusion model dataset.} Models here are trained on the 4-class ProGAN data. We transpose the table for better readability. Notations are consistent with Table~\ref{tab:table1}.}
\resizebox{\textwidth}{!}{
\begin{tabular}{cl|cccccccc|c}
\multicolumn{2}{c|}{Dataset} & Wang \cite{wang2020cnn} & Durall \cite{durall2020watch} & Frank \cite{frank2020leveraging} & PatchFor \cite{chai2020makes} & F3Net \cite{qian2020thinking} & Blend$\dagger$ \cite{shiohara2022detecting} & LGrad \cite{tan2023learning} & UniFD \cite{ojha2023towards} & Ours \\
\shline
\multicolumn{2}{c|}{PNDM}        & $50.8$ / $90.3$ & $44.5$ / $47.3$ & $44.0$ / $38.2$ & $50.2$ / $99.9$ & $72.8$ / $99.5$ & $48.2$ / $48.1$ & $69.8$ / $98.5$ & $75.3$ / $92.5$ & $\bf 99.3$ / $\bf 100.0$ \\
\multicolumn{2}{c|}{Guided} & $54.9$ / $66.6$ & $40.6$ / $42.3$ & $53.4$ / $52.5$ & $74.2$ / $81.4$ & $69.2$ / $70.8$ & $58.3$ / $63.4$ & $\bf 86.6$ / $\bf 100.0$ & $75.7$ / $85.1$ & $76.1$ / $92.0$ \\
\multicolumn{2}{c|}{DALL-E}        & $51.8$ / $61.3$ & $55.9$ / $58.0$ & $57.0$ / $62.5$ & $79.8$ / $99.1$ & $71.6$ / $79.9$ & $52.4$ / $51.6$ & $88.5$ / $97.3 $ & $89.5$ / $96.8$ & $\bf 98.8$ / $\bf 99.8$ \\ 
\multicolumn{2}{c|}{VQ-Diffusion} & $50.0$ / $71.0$ & $38.6$ / $38.3$ & $51.7$ / $66.7$ & $100.0$ / $100.0$&$100.0$ / $100.0$& $77.1$ / $82.6$ & $96.3$ / $100.0$& $83.5$ / $97.7$ & $\bf 100.0$ / $\bf 100.0$ \\ \cline{1-2}
\multirow{3}{*}{LDM} & 200 steps  & $52.0$ / $64.5$ & $61.7$ / $61.7$ & $56.4$ / $50.9$ & $95.6$ / $\bf 99.9$ & $73.4$ / $83.3$ & $52.6$ / $51.9$ & $94.2$ / $99.1$ & $90.2$ / $97.1$ & $\bf 98.6$ / $99.8$ \\
                                            & 200 w/ CFG & $51.6$ / $63.1$ & $58.4$ / $58.5$ & $56.5$ / $52.1$ & $94.0$ / $\bf 99.8$ & $80.7$ / $89.1$ & $51.9$ / $52.6$ & $\bf 95.9$ / $99.2$ & $77.3$ / $88.6$ & $94.9$ / $99.1$ \\
                                            & 100 steps  & $51.9$ / $63.7$ & $62.0$ / $62.6$ & $56.6$ / $51.3$ & $95.8$ / $99.8$ & $74.1$ / $84.0$ & $53.0$ / $54.0$ & $94.8$ / $99.2$ & $90.5$ / $97.0$ & $\bf 98.7$ / $\bf 99.9$ \\ \cline{1-2}
\multirow{3}{*}{Glide} & 100-27   & $53.0$ / $71.3$ & $48.9$ / $46.9$ & $50.4$ / $40.8$ & $82.8$ / $99.1$ & $87.0$ / $94.5$ & $59.4$ / $64.1$ & $87.4$ / $93.2$ & $90.7$ / $97.2$ & $\bf 94.4$ / $\bf 99.1$ \\
                                              & 50-27    & $54.2$ / $76.0$ & $51.7$ / $49.9$ & $52.0$ / $42.3$ & $84.9$ / $98.8$ & $88.5$ / $95.4$ & $64.2$ / $68.3$ & $90.7$ / $95.1$ & $91.1$ / $97.4$ & $\bf 94.7$ / $\bf 99.4$ \\
                                              & 100-10   & $53.3$ / $72.9$ & $54.9$ / $52.3$ & $53.6$ / $44.3$ & $87.3$ / $\bf 99.7$ & $88.3$ / $95.4$ & $58.8$ / $63.2$ & $89.4$ / $94.9$ & $90.1$ / $97.0$ & $\bf 94.2$ / $99.2$ \\
\hline
\multicolumn{2}{c|}{\cellcolor{lightgray!20}{Mean}}        & \cellcolor{lightgray!20}{$52.4$ / $70.1$} & \cellcolor{lightgray!20}{$51.7$ / $51.8$} & \cellcolor{lightgray!20}{$53.2$ / $50.2$} & \cellcolor{lightgray!20}{$84.5$ / $97.8$} & \cellcolor{lightgray!20}{$80.6$ / $89.2$} & \cellcolor{lightgray!20}{$57.6$ / $60.0$} & \cellcolor{lightgray!20}{$89.4$ / $97.7$} & \cellcolor{lightgray!20}{$85.4$ / $94.6$} & \cellcolor{lightgray!20}{$\bf 95.0$ / $\bf 98.8$} \\
\end{tabular}}
\label{tab:table2}
\vspace{-3mm}
\end{table*}

\noindent \textbf{Comparisons on GANs dataset.}
Table~\ref{tab:table1} reports the comparisons on the GANs dataset \cite{wang2020cnn} with two different training data settings. Results show that our FatFormer consistently exceeds pre-trained-based LGrad \cite{tan2023learning} and UniFD \cite{ojha2023towards}. Specifically, under 4-class supervision, FatFormer outperforms the current state-of-the-art method UniFD by a significant $9.3\%$ ACC and $1.4\%$ AP with the same pre-trained CLIP model, achieving $98.4\%$ ACC and $99.7\%$ AP. Besides, for the other 2-class supervision setting, similar trends are observed with the ones under 4-class supervision, when compared with pre-trained-based methods. Moreover, we also compare FatFormer with representative image-based \cite{wang2020cnn,chai2020makes,shiohara2022detecting} and frequency-based methods \cite{durall2020watch,frank2020leveraging,qian2020thinking,jeong2022bihpf,jeong2022frepgan} in Table~\ref{tab:table1}. Our approach can also easily outperform all of them with a larger improvement.

The above evidence indicates the necessity of forgery adaptation for pre-trained models. Beyond the impressive performance, more importantly, our FatFormer provides an effective paradigm of how to incorporate pre-trained models in the synthetic image detection task.

\begin{table*}[t]
\centering
\captionsetup{font=small}
\caption{\textbf{Ablation experiments for FatFormer.} Evaluated on GANs dataset. Default settings are marked in \colorbox{lightgray!20}{gray}.}
\vspace{-2mm}
    \begin{subtable}[t]{0.36\textwidth}
        \centering
        \renewcommand\arraystretch{1.0}
        \caption{\textbf{Forgery-aware adapter implementations.} In the proposed framework, both image (img) and frequency (freq) domains are essential for building generalized forgery representation.}
        \resizebox{0.97\linewidth}{!}{
        \begin{tabular}{cc|cc} 
        w/ img domain & w/ freq domain & ACC$_M$ & AP$_M$ \\
        \shline
        \cellcolor{lightgray!20}{\checkmark} & \cellcolor{lightgray!20}{\checkmark} & \cellcolor{lightgray!20}{$\bf 98.4 $} & \cellcolor{lightgray!20}{$\bf 99.7$} \\
        \checkmark & $\times$   & $ 95.4 $ & $ 99.6 $ \\
        $\times$   & \checkmark & $ 97.3 $ & $ 99.6 $ \\
        \end{tabular}}
        \label{tab:ablationadapterimplementation}
    \end{subtable}
    \hfill
    \begin{subtable}[t]{0.25\textwidth}
        \centering
        \renewcommand\arraystretch{1.0}
        \caption{\textbf{Frequency band interactions.} Both inter- and intra-band attentions are important for modeling forgery traces in the frequency domain. }
        \resizebox{0.88\linewidth}{!}{
        \begin{tabular}{c|cc}
        interaction & ACC$_{M}$ & AP$_{M}$ \\
        \shline
        intra & $ 97.4 $ & $ 99.7 $ \\
        inter & $ 96.6 $ & $ 99.6 $ \\
        \cellcolor{lightgray!20}{intra \& inter} & \cellcolor{lightgray!20}{$\bf 98.4 $} & \cellcolor{lightgray!20}{$\bf 99.7$}
        \end{tabular}}
        \label{tab:ablationfreq}
    \end{subtable}
    \hfill
    \begin{subtable}[t]{0.35\textwidth}
        \centering
        \renewcommand\arraystretch{1.0}
        \caption{\textbf{Benefits of supervision in vision-language space.} On the model only with img input, text is first added for building contrastive (contra) objectives. Then, we apply the proposed augmented (aug) contra strategy.}
        \resizebox{0.95\linewidth}{!}{
        \begin{tabular}{cc|cc}
        input modality & strategy & ACC$_M$ & AP$_{M}$ \\
        \shline
        only img & linear probe      & $ 95.3 $ & $ 99.2 $ \\
        img $\&$ text & contra & $ 96.4 $ & $ 99.6 $ \\
        \cellcolor{lightgray!20}{img $\&$ text} & \cellcolor{lightgray!20}{aug contra} & \cellcolor{lightgray!20}{$\bf 98.4 $} & \cellcolor{lightgray!20}{$\bf 99.7$} \\
        \end{tabular}
        }
        \label{tab:ablationvisonlanguagespace}
     \end{subtable}
     \hfill
     \begin{subtable}[t]{0.42\textwidth}
        \centering
        \renewcommand\arraystretch{1.0}
        \vspace{2mm}
        \caption{\textbf{Text prompt designs.} The auto embedding and img condition can benefit the performance, especially by considering the correlation between prompt and img patch tokens.}
        \resizebox{0.88\linewidth}{!}{
        \begin{tabular}{cc|cc}
        prompt designs & w/ img condition & ACC$_{M}$ & AP$_{M}$ \\
        \shline
        fixed template  & $\times$  & $ 95.5 $ & $ 99.6 $ \\
        auto embedding  & $\times$  & $ 96.4 $ & $ 99.6 $ \\
        auto embedding  & CLS token & $ 98.1 $ & $ 99.7 $ \\
        \cellcolor{lightgray!20}{auto embedding} & \cellcolor{lightgray!20}{patch tokens} & \cellcolor{lightgray!20}{$\bf 98.4 $} & \cellcolor{lightgray!20}{$\bf 99.7 $} \\
        \end{tabular}}
        \label{tab:ablationpromptdesign}
     \end{subtable}
     \hfill
    \begin{subtable}[t]{0.53\textwidth}
        \centering
        \renewcommand\arraystretch{1.0}
        \vspace{2mm}
        \caption{\textbf{Model components.} Both components are essential in our FatFormer. We also conduct an extra experiment to test the zero-shot performance by removing the forgery-aware adapter and language-guided alignment.}
        \resizebox{0.9\linewidth}{!}{
        \begin{tabular}{c|cc}
        module components & ACC$_{M}$ & AP$_{M}$ \\
        \shline
        none for zero-shot      & $ 66.6 $ & $ 74.3 $ \\
        forgery-aware adapter   & $ 95.3 $ & $ 99.2 $ \\
        language-guided alignment & $ 91.5 $ & $ 98.1 $ \\
        \cellcolor{lightgray!20}{forgery-aware adapter $\&$ language-guided alignment} & \cellcolor{lightgray!20}{$\bf 98.4 $} & \cellcolor{lightgray!20}{$\bf 99.7$} \\
        \end{tabular}}
        \label{tab:ablationmodelcomponnent}
     \end{subtable}
\vspace{-3mm}
\end{table*}

\noindent \textbf{Comparisons on diffusion model dataset.}
To further demonstrate the effectiveness of FatFormer, we provide comparisons with existing detection methods on the diffusion model dataset \cite{ojha2023towards}. The results are shown in Table~\ref{tab:table2}. Note that all the compared methods are trained on 4-class ProGAN data. This test setting is more challenging as forged images are created by various diffusion models with completely different generating theories and processes from GANs.
Surprisingly, FatFormer generalizes well for diffusion models, achieving $95.0\%$ ACC and $98.8\%$ AP. 

Compared with pre-trained-based LGrad \cite{tan2023learning} and UniFD \cite{ojha2023towards}, FatFormer also works better than both of them when handling diffusion models. For example, our approach surpasses UniFD by $9.6\%$ ACC and $4.2\%$ AP. Moreover, we find that even with powerful CLIP as the pre-trained model, UniFD only achieves a similar result (about $85\%$ ACC) like PatchFor \cite{chai2020makes}. We argue this is mainly because the fixed pre-trained paradigm is prone to yield detectors with insufficient learning regarding forgery artifacts. Thus, our FatFormer, which presents an adaptive transformer framework with forgery adaptation and reasonable contrastive objectives, can achieve much better results.

\subsection{Ablation study}
\label{sec:ablation}
We conduct several ablation experiments to verify the effectiveness of key elements in our FatFormer. Unless specified, we report the mean of accuracy (ACC$_M$) and average precision (AP$_M$) on the GANs dataset under the training setting of 4-class ProGAN data.

\noindent \textbf{Forgery-aware adapter implementations.}
We ablate the effects of considering the image domain and frequency domain in the forgery-aware adapter. The results are shown in Table~\ref{tab:ablationadapterimplementation}. We observe severe performance degradation when removing either of these two domains, especially for the frequency domain with over $-3.0\%$ ACC gaps. We conclude that both image and frequency domains are essential in FatFormer for synthetic image detection. The image forgery extractor collects the local low-level forgery artifacts, {\em e.g.}, blur textures, while the frequency forgery extractor explores and gathers the forgery clues among different frequency bands, together building a comprehensive local viewpoint for the adaptation of image features. 

For the frequency forgery extractor, both interactions built by inter-band and intra-band attentions are important in our FatFormer. Table~\ref{tab:ablationfreq} shows the ablation.

\noindent \textbf{Benefits of supervision in vision-language space.}
Table~\ref{tab:ablationvisonlanguagespace} provides the comparisons between different supervising strategies for FatFormer, including (i) linear probing with image modality, (ii) vanilla contrastive objectives between image CLS token and text prompt embeddings, which masked out the text-guided interactor, and (iii) our augmented contrastive objectives. The results demonstrate that introducing text prompts for contrastive supervision benefits the generalization of detection. 
We conjecture this is mainly because CLIP provides a stable alignment between real image and text representation with pre-training, thus yielding a mismatching when handling a fake image with text prompts. As potential evidence, we find that only adopting LGA can still achieve an accuracy of $91.5\%$ ACC (Table~\ref{tab:ablationmodelcomponnent}). 
Besides, we observe that the proposed augmented contrastive objectives can further boost generalizability by directing the image encoder to concentrate on forgery-related representations, bringing a $2.0\%$ ACC gain over the vanilla implementation.

\noindent \textbf{Text prompt designs.}
Table~\ref{tab:ablationpromptdesign} gives the results of constructing the text prompt with different prompt designs and image conditions. The results validate that both auto context embeddings and image conditions are important in text prompt designs. Compared with using a fixed hand-crafted template, {\em e.g.}, `this photo is', the design of auto context embedding improves by $0.9\%$ ACC, due to its abstract exploration in word embedding spaces.
Besides, it is better to adopt image patch tokens as conditions to enhance these auto context embeddings, containing more local context details, rather than the global image CLS token.

\begin{figure*}[t]
\centering
\includegraphics[width=0.95\linewidth]{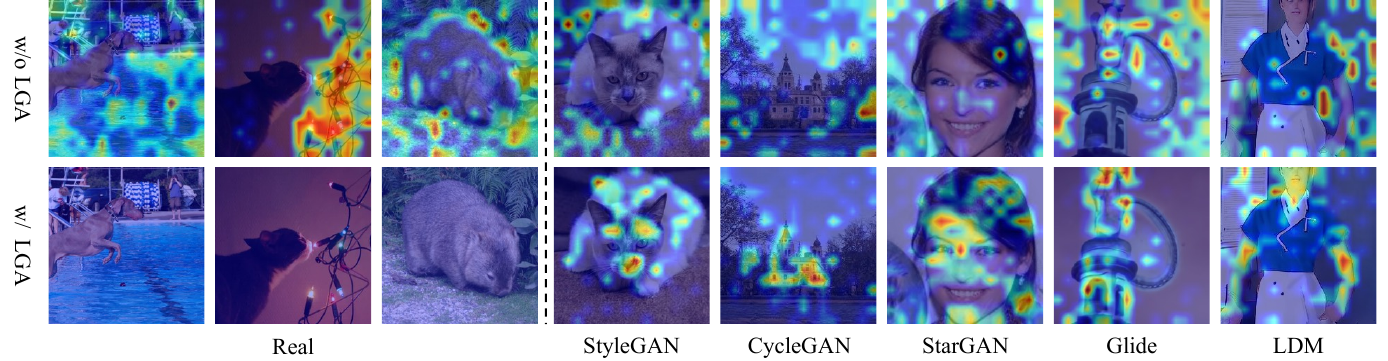}
\caption{\textbf{Comparison of model attention with (w/ ) and without (w/o) language-guided alignment.} We visualize the gradient norm of FatFormer (second row) and FatFormer without language-guided alignment (first row) by \cite{selvaraju2017grad}. FatFormer provides more responses among semantic foreground patches in fake images, while almost no response for real ones. The salient region is visualized by bright color.}  
\label{fig:figure4}
\vspace{-4mm}
\end{figure*}

\noindent \textbf{Model components.}
Tabel~\ref{tab:ablationmodelcomponnent} gives the ablation of two proposed model components, {\em i.e.}, forgery-aware adapter and language-guided alignment. Large performance drops ($-6.9\%$ ACC and $-1.6\%$ AP) are observed when adopting the previous fixed pre-trained paradigm by removing the forgery-aware adapter. This explains the necessity of forgery adaptation of pre-trained models. On the other hand, the proposed language-guided alignment, which considers the augmented contrastive objectives in the vision-language space, also provides better supervision for the forgery adaptation than simply adopting binary labels, bringing $3.1\%$ ACC and $0.5\%$ AP gains. 

As shown in Figure~\ref{fig:figure4}, using language-guided alignment obtains more concentration on semantic foreground patches, where anomalies, {\em e.g.}, unrealistic objects, textures, or structures often occur. Therefore, our FatFormer can obtain generalized forgery representations by focusing on local forgery details, resulting in the improvement of the generalizability of synthetic image detection.

\subsection{More analysis}
Here, we analyze our FatFormer on different architectures and pre-training strategies.

\noindent \textbf{Analysis on different architectures.}
While FatFormer is constructed upon the identical CLIP framework \cite{radford2021learning} as employed in UniFD \cite{ojha2023towards}, the proposed forgery adaptation strategy is transferrable to alternative architectures.
Presented in the upper section of Table~\ref{tab:moreanalysis} are the ACC$_M$ and AP$_M$ scores for four distinct architectures, including two variations of multi-modal structures pre-trained by CLIP and two variants of image-based Swin transformer \cite{liu2021Swin} pre-trained on ImageNet 22k \cite{deng2009imagenet}.
The comparisons between models with and without FatFormer verify the efficacy of integrating forgery adaptation among different pre-trained architectures, significantly facilitating the performance of synthetic image detection.

\noindent \textbf{Analysis on different pre-training strategies.} 
We further conduct an assessment of the efficacy of forgery adaptation across models employing different pre-training strategies. 
Utilizing ViT-L \cite{dosovitskiy2020image} as the baseline, we validate two well-known pre-training approaches: MAE \cite{he2022masked} and CAE \cite{chen2023context}.
The evaluations are shown in the lower segment of Table~\ref{tab:moreanalysis}. 
We observe that incorporating the forgery adaptation in our FatFormer can lead to a consistent increase in performance across diverse pre-training strategies, demonstrating the robustness and transferability of our approach.

\begin{table}[t]
\centering
\setlength{\tabcolsep}{5pt}
\renewcommand{\arraystretch}{1.4}
\footnotesize
\captionsetup{font=small}
\caption{\textbf{Analysis on different architectures and pre-training strategies.} Beyond UniFD, the forgery adaptation in FatFormer can also consistently boost various architectures and different pre-training strategies. We report the mean of ACC and AP (in the formulation of ACC$_M$ / AP$_M$) on both GANs and diffusion model (DMs) datasets. `IN-22K'= ImageNet 22k.}
\resizebox{\linewidth}{!}{
\begin{tabular}{c|c|c|ccc}
Architecture & Pre-training & w/ Ours & GANs & DMs \\
\shline
\multirow{2}{*}{\thead{ViT-B/16\\Text-512}} & \multirow{2}{*}{CLIP \cite{radford2021learning}}
  & $\times$   & $ 83.8 $ / $ 94.4 $ & $ 77.2 $ / $ 91.1 $ \\ 
& & \checkmark & $\bf 95.3 $ / $\bf 99.5 $ & $\bf 91.6 $ / $\bf 97.8 $ \\ 
\hline
\multirow{2}{*}{\thead{ViT-L/14\\Text-768}} & \multirow{2}{*}{CLIP \cite{radford2021learning}}
  & $\times$   & $ 89.1 $ / $ 98.3 $ & $ 85.4 $ / $ 94.6 $ \\ 
& & \checkmark & $\bf 98.4 $ / $\bf 99.7 $ & $\bf 95.0 $ / $\bf 98.8 $ \\
\hline
\multirow{2}{*}{Swin-B} & \multirow{2}{*}{IN-22K \cite{deng2009imagenet}}
  & $\times$   & $ 82.5 $ / $ 93.8 $ & $ 72.2 $ / $ 88.8 $ \\ 
& & \checkmark & $\bf 89.6 $ / $\bf 98.2 $ & $\bf 76.1 $ / $\bf 96.1 $ \\ 
\hline
\multirow{2}{*}{Swin-L} & \multirow{2}{*}{IN-22K \cite{deng2009imagenet}}
  & $\times$   & $ 86.4 $ / $ 95.7 $ & $ 74.4 $ / $ 90.8 $ \\ 
& & \checkmark & $\bf 90.7 $ / $\bf 98.4 $ & $\bf 79.3 $ / $\bf 96.7 $ \\ 
\shline
\multirow{4}{*}{ViT-L/16} & \multirow{2}{*}{MAE \cite{he2022masked}}
  & $\times$   & $ 75.7 $ / $ 92.8 $ & $ 70.9 $ / $ 92.3 $ \\ 
& & \checkmark & $\bf 85.2 $ / $\bf 96.7 $ & $\bf 88.5 $ / $\bf 98.4 $ \\ 
\cline{2-5}
& \multirow{2}{*}{CAE \cite{chen2023context}}
  & $\times$   & $ 76.1 $ / $ 95.9 $ & $ 64.9 $ / $ 91.7 $ \\ 
& & \checkmark & $\bf 88.1 $ / $\bf 98.0 $ & $\bf 76.1 $ / $\bf 96.2 $ \\ 
\end{tabular}}
\label{tab:moreanalysis}
\vspace{-4mm}
\end{table}

\section{Conclusion}
In this paper, we present a novel adaptive transformer, FatFormer, for generalizable synthetic image detection. With two core designs, including the forgery-aware adapter and language-guided alignment, for the forgery adaption of pre-trained models, the proposed approach outperforms the previous fixed pre-trained paradigm by a large margin. Besides, the forgery adaption in FatFormer is also flexible, which can be applied in various pre-trained architectures with different pre-training strategies. We hope FatFormer can provide insights for exploring better utilization of pre-trained models in the synthetic image detection field.

\noindent \textbf{Limitations and future works.} 
FatFormer generalizes well on most generative methods, while we still have space to improve in diffusion models, {\em e.g.}, Guided \cite{dhariwal2021diffusion}. 
Elucidating the distinctions and associations among images produced by diffusion models and GANs is needed to build stronger forgery detectors.
The investigation of this problem is left in future work. 
Besides, how to construct a better pretext task special for synthetic image detection in pre-training is also worth a deeper study.

{
    \small
    \bibliographystyle{ieeenat_fullname}
    \bibliography{main}
}

\clearpage
\begin{appendices}
\section{Appendix}
In this appendix, we first discuss the \textit{potential negative societal impacts} (refer to Section~\ref{impacts}) that may arise in practical scenarios. Then, an in-depth exploration of \textit{ablation studies} (explicated in Section~\ref{ablation}) is presented, delineating the influence of hyper-parameters employed within our approach. Lastly, a comprehensive analysis is conducted to assess the efficacy of forgery adaptation in enhancing \textit{robustness} (outlined in Section~\ref{robustness}) against image perturbations.

\subsection{Broader impacts}
\label{impacts}
The development of synthetic image detection tools, while aiming to combat misinformation, may lead to unintended consequences in content moderation. Legitimate content that exhibits characteristics similar to forgeries may be mistakenly flagged, impacting normal information (based on image modality) sharing. These issues need further research and consideration when deploying this work to practical applications for content moderation.

\subsection{More Ablations}
\label{ablation}
We provide more ablation studies on the hyper-parameters used in our FatFormer. The training and evaluating settings are the same as Section~\textcolor{red}{4.3}.

\noindent \textbf{Number of auto context embeddings.}
FatFormer combines the enhanced context embeddings and [CLASS] embeddings to construct the set of possible text prompts. Here, we ablate the effects of how a pre-defined number of context embeddings in text prompts affects the performance in the following table:
\begin{table}[h]
\centering
\vspace{-3mm}
\renewcommand\arraystretch{1.0}
\resizebox{0.51\linewidth}{!}{
\begin{tabular}{c|cc}
\#embeddings & ACC$_{M}$ & AP$_{M}$ \\
\shline
$ 4 $  & $ 97.6 $ & $ 99.0 $ \\
\cellcolor{lightgray!20}{$ 8 $} & \cellcolor{lightgray!20}{$\bf 98.4 $} & \cellcolor{lightgray!20}{$\bf 99.7$} \\
$ 16 $ & $ 97.8 $ & $ 99.6 $ \\
\end{tabular}}
\vspace{-3mm}
\end{table}

\noindent One can see that $8$ auto context embeddings are good enough and achieve better results than $16$ embeddings. Thus, we set the number as $8$ by default in this paper. 

\noindent \textbf{Number of forgery-aware adapters.}
To achieve effective forgery adaptation, FatFormer develops the forgery-aware adapter and integrates it with the ViT image encoder. The number of inserted forgery-aware adapters is to be explored. The following table lists the relevant ablations:
\begin{table}[h]
\centering
\vspace{-3mm}
\renewcommand\arraystretch{1.0}
\resizebox{0.46\linewidth}{!}{
\begin{tabular}{c|cc}
\#adapters & ACC$_{M}$ & AP$_{M}$ \\
\shline
$ 2 $ & $ 97.2 $ & $ 99.6 $ \\
\cellcolor{lightgray!20}{$ 3 $} & \cellcolor{lightgray!20}{$\bf 98.4 $} & \cellcolor{lightgray!20}{$\bf 99.7$} \\
$ 4 $ & $ 96.5 $ & $ 99.7 $ \\
\end{tabular}}
\vspace{-3mm}
\end{table}

\noindent We observe that inserting $3$ forgery-aware adapters in the image encoder is able to achieve good performance. Therefore, we set $3$ as the default number of the forgery-aware adapter in our FatFormer.

\noindent \textbf{Kernel size of image forgery extractor.} To capture low-level image artifacts, we introduce a lightweight image forgery extractor in the proposed forgery-aware adapter, including two convolutional layers and a ReLU. We also explore settings of the kernel size of convolutional layers, as follows:

\begin{table}[h]
\centering
\vspace{-3mm}
\renewcommand\arraystretch{1.0}
\resizebox{0.46\linewidth}{!}{
\begin{tabular}{c|cc}
kernel size & ACC$_{M}$ & AP$_{M}$ \\
\shline
\cellcolor{lightgray!20}{$ 1 $} & \cellcolor{lightgray!20}{$\bf 98.4 $} & \cellcolor{lightgray!20}{$\bf 99.7$} \\
$ 3 $ & $ 96.4 $ & $ 99.7 $ \\
$ 5 $ & $ 95.6 $ & $ 99.6 $ \\
\end{tabular}}
\vspace{-3mm}
\end{table}

\noindent We find that using $1\times 1$ kernel yields superior results in constructing the image forgery extractor. We conjecture that this is mainly because the intermediate image patch tokens in ViT encode high-level semantic information of different image patches, which may not provide useful low-level similarity among adjacent positions like the ones in traditional convolutional networks. Thus, larger kernels, designed to fuse adjacent patch tokens, may introduce disturbance to the modeling process of ViT and damage the performance.

\begin{figure*}[h]
\centering
\begin{tikzpicture}[font=\footnotesize]
\begin{axis}[
    width=\textwidth,
    height=8cm,
    ymajorgrids,
    yminorgrids,
    grid style={line width=.1pt, draw=gray!20},
    major grid style={line width=.2pt,draw=gray!50},
    minor y tick num=3,
    ybar,
    enlargelimits=0.08,
    legend style={
    at={(0.93,0.98)},
      anchor=north,legend columns=1,font=\scriptsize},
    legend image code/.code={
        \draw [#1] (0cm,-0.1cm) rectangle (0.2cm,0.25cm); },
    ylabel={Accuracy (\%)},
    y label style={at={(0.01,0.5)}},
    symbolic x coords={ProGAN,StyleGAN,StyleGAN2,BigGAN,CycleGAN,StarGAN,GuaGAN,Deepfake,Mean},
    xtick=data,
    nodes near coords,
    nodes near coords align={vertical},
    every node near coord/.append style={
        /pgf/number format/fixed zerofill,
        /pgf/number format/precision=1}
    ]
\addplot[bar width=0.6cm,fill=selfblue!50] coordinates {(ProGAN,95.6) (StyleGAN,85.3) (StyleGAN2,80.5) (BigGAN,85.5) (CycleGAN,84.4) (StarGAN,84.4) (GuaGAN,87.5) (Deepfake,74.8) (Mean,84.7)};
\addlegendentry{UniFD}
\addplot[bar width=0.6cm,fill=selfgreen!50] coordinates {(ProGAN,97.0) (StyleGAN,90.4) (StyleGAN2,89.0) (BigGAN,93.7) (CycleGAN,94.8) (StarGAN,98.2) (GuaGAN,95.7) (Deepfake,88.1) (Mean,93.4)};
\addlegendentry{Ours}
\end{axis}
\end{tikzpicture}
\caption{\textbf{Robustness comparisons with combined four image perturbations.} We report the accuracy results on the GANs dataset. By considering the forgery adaptation, our FatFormer works better on all generative models than UniFD which adopts the fixed pre-trained paradigm.}
\vspace{-3mm}
\label{fig:figure5}
\vspace{15cm}
\end{figure*}
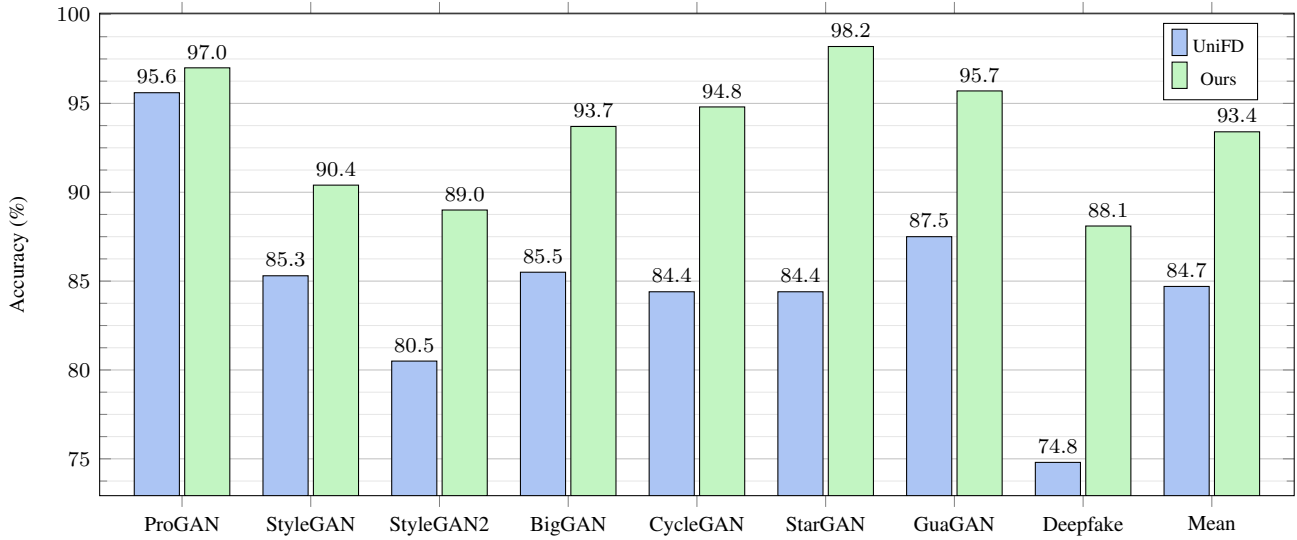

\subsection{Robustness on image perturbation}
\label{robustness}
To evaluate the effects of forgery adaptation in FatFormer on robustness, we apply several common image perturbations to the test images, following [\textcolor{cvprblue}{12}, \textcolor{cvprblue}{46}]. Specifically, we adopt random cropping, Gaussian blurring, JPEG compression, and Gaussian noising, each with a probability of 50\%. The detailed perturbation configures can be found in [\textcolor{cvprblue}{12}]. Based on the GANs dataset, we compare our FatFormer with UniFD [\textcolor{cvprblue}{35}] and LGrad [\textcolor{cvprblue}{46}], which adopts the fixed pre-trained paradigm. The results are shown in the following table:
\begin{table}[h]
\centering
\vspace{-3mm}
\renewcommand\arraystretch{1.15}
\resizebox{0.8\linewidth}{!}{
\begin{tabular}{ll|cc}
Perturbation & Method & ACC$_{M}$ & AP$_{M}$ \\
\shline
\multirow{2}{*}{Gaussian blurring} 
& LGrad & $ 78.5 $ & $ 83.2 $\\
& UniFD & $ 78.1 $ & $ 93.0 $ \\
& FatFormer & $\bf 90.7 $ & $\bf 98.1 $ \\
\hline
\multirow{2}{*}{random cropping} 
& LGrad & $ 85.0 $ & $ 91.9 $ \\
& UniFD & $ 88.9 $ & $ 98.1 $ \\
& FatFormer & $\bf 98.2 $ & $\bf 99.7 $ \\
\hline
\multirow{2}{*}{JPEG compression} 
& LGrad & $ 69.5 $ & $ 81.2 $ \\
& UniFD & $ 88.4 $ & $ 97.7 $ \\
& FatFormer & $\bf 95.9 $ & $\bf 99.2 $ \\
\hline
\multirow{2}{*}{Gaussian noising} 
& LGrad & $ 69.1 $ & $ 79.4 $ \\
& UniFD & $ 82.6 $ & $ 93.9 $ \\
& FatFormer & $\bf 88.0 $ & $\bf 96.5 $ \\
\end{tabular}}
\vspace{-3mm}
\end{table}

\noindent It can be observed that our approach exceeds UniFD by a larger margin, {\em e.g.}, over $+12.0\%$ facing Gaussian blurring. This is mainly because FatFormer obtains well-generalized forgery representations with the proposed forgery adaption, as analyzed in Section~\textcolor{red}{4.3}.

Moreover, we also consider a more real-world scenario by combining all four types of perturbation. The results are illustrated in Figure~\ref{fig:figure5}. Compared with UniFD, our FatFormer also beats it on all testing GAN methods, further suggesting the robustness improvement brought by forgery adaptation.
\end{appendices}

\end{document}